# PartsNet: A Unified Deep Network for Automotive Engine Precision Parts Defect Detection

Zhenshen Qu[1], Jianxiong Shen[2], Ruikun Li[1], Junyu Liu[1], and Qiuyu Guan[1]
[1]Department of Control Science and Engineering, HIT, Harbin, China
[2]State Key Laboratory of advanced welding and joining, Harbin Institute of Technology, Harbin, China
Email: miraland@hit.edu.cn, 15549298924@163.com

## ABSTRACT

Defect detection is a basic and essential task in automatic parts production, especially for automotive engine precision parts. In this paper, we propose a new idea to construct a deep convolutional network combining related knowledge of feature processing and the representation ability of deep learning. Our algorithm consists of a pixel-wise segmentation Deep Neural Network (DNN) and a feature refining network. The fully convolutional DNN is presented to learn basic features of parts defects. After that, several typical traditional methods which are used to refine the segmentation results are transformed into convolutional manners and integrated. We assemble these methods as a shallow network with fixed weights and empirical thresholds. These thresholds are then released to enhance its adaptation ability and realize end-to-end training. Testing results on different datasets show that the proposed method has good portability and outperforms the state-of-the-art algorithms.

## 1. INTRODUCTION

Visual defects, such as scratches and cracks, can be widely observed on finished industrial products, which put a major threat to production quality. Some of the defects for mechanical parts are shown in Figure 1 Therefore, precise and reliable automatic defect detection of visual defects instead of manual work is a fundamental task in the manufacturing process.

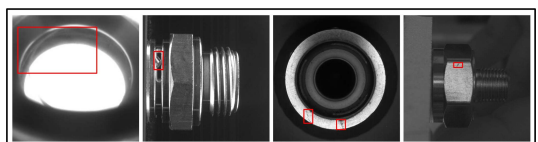

**Figure 1. Different types of defects in precision parts in industrial production.**

By now, various image processing techniques for defect detection have been widely investigated, such as thresholding [1], segmentation-based method [2], edge detection based method [4]. However, due to the diversity of defects' geometric features, which may be completely different among detected products, most existing methods can only be used for a few kinds of specific products or a batch of products. Along with the development of feature extraction techniques, feature based methods are widely used to detect defects [5]. These methods utilize hand-designed features from domain knowledge [8] and can be divided into three main types: statistical, filter based and model based. As for statistical methods, Wang et al. [9] proposed a novel adaptive artificial fish swarm algorithm optimized Hough transform (AAFSA-HT) to detect the circular parts automatically. Reinhold et al. [10] presented a statistical approach to feature selection, classification, detection and fusion applied to high-resolution analysis of rail head images. Filter based approaches include spatial domain and frequency domain filtering design [11]. A.S. et al. [12] extracted defects features from the Log-Gabor filter bank response in homogeneous flat surface products. For model-based methods, K et al. [13] proposed a new Haar-Weibull-variance (HWV) model for steel surface defect detection in an unsupervised manner. In conclusion, the effects of these algorithms, in which most these features are fully hand-crafted, are largely impacted by designer's subjective experience and cognition. That means that these specially designed features are just appropriate for specific products and will show weak adaptation for other products.

Since 2012, deep learning has developed rapidly and achieved a great success in various fields such as VGGNet [14] for image recognition, FCN [15] for segmentation and Faster R-CNN [16], SSD [17] as well as YOLO [18] for object detection. Deep learning methods, well-known for their ability to learn from data automatically and realize end-to-end control, are gradually applied in the field of defect detection in recent years. But, in-depth scientific research reports in this domain are limited although many detection algorithms have been presented in companies. F. S. Roohi et al. [19] presented an application of deep convolutional neural networks (DCNNs) for automatic classification of rail surface defects. This network can recognize low-resolution images, but it is only suitable for normative and apparent defects. Yu et al. [20] proposed a two-stage fully convolutional networks for surface defect inspection in industrial environment. However, straight outputs of fully convolutional networks are not accurate enough and may exist small-area false detection, which can heavily limit classification performance. It seems that some additional steps and proper criteria on classification are crucial to improve algorithm effect.

In this paper, a new method is proposed for parts defect detection and network design combining traditional image processing and deep learning. The complete structure is composed of pixel-wise segmentation network and refining network. We demonstrate that the refining pipeline consisting of density slicing, region segmentation and area filtration is equivalent to a simple network with empirical thresholds. As these thresholds are fixed, their adaptability is limited and they cannot distinguish true defects from complex background accurately. Inspired by [21], the simple network is then transformed into a convolutional network with some extra layers to enhance its adaptability, and thresholds are released to learn classification rules adaptively from data. The specially designed network for parts defects is called PartsNet. Benefiting from our method, the false detection rate and the missing detection rate have significantly decreased. Finally, we test our network on five different parts defects and achieve reliable results based on a small-sized dataset.

The key contributions of this paper are as follows:

1. A new method to biases design in network construction and refine pixel-wise segmentation results using traditional knowledge of feature processing.

2. A novel network for parts defects called PartsNet is proposed. Typical feature processing method for parts defects including density slicing, region segmentation and area filtration are transformed into an equivalent network which is then integrated into a unified network with a deep learning network and achieves end-to-end training.

3. A reusable network achieves improved adaptation on different types of parts defects automatically without any fine tuning.

## 2. PROPOSED NETWORK

### 2.1 Method Overview

In this work, the main idea is to build an efficient network for parts defect detection, which integrates the representation ability of deep learning and the advantages of traditional methods. Based on this thought, our proposed network is composed of two parts. Firstly, we use a fully convolutional network as our segmentation network to learn basic features of defects. Then, a refining network, aimed to filter out the small-area false detection and get correct classification output, will be applied on the segmentation results. The refining steps include density slicing, region segmentation and area filtration. In particular, all these refining operations are transformed into a convolutional network and participate in the training process.

The whole system is shown in Figure 2.

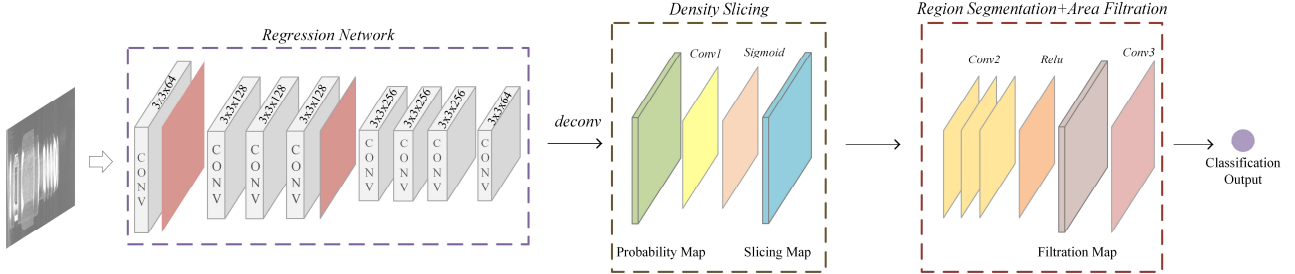

**Figure 2. Structure of the proposed network for defect detection.**

### 2.2 Segmentation Network

The introduction mentioned that the classification network cannot meet the requirements of defect detection. For a classification network, it is hard to learn accurate defect features from complex background due to the fuzzy edge. In contrast, the input labels of segmentation network are marked on the pixel scale, including the location and size of defects. This simplifies the learning process of model and benefits edge feature extraction.

Traditional classification networks utilize the fully connected layers to obtain the global information of images. In order to obtain the pixel-wise predicted intensity value, we changed the last several layers of full connection layer into the full convolution layer, which is widely used in the field of image segmentation [15]. That also brings about two advantages: firstly, it can accept input images of any size; secondly, the number of parameters is greatly reduced and it is more efficient.

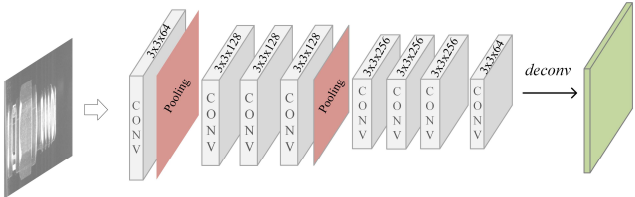

**Figure 3. Structure of the segmentation network.**

As shown in Figure 3, the first 2 blocks are of the same Conv-BN-ReLU-Pooling structure used for initial edge learning. According to the results of [14], under the condition of constant receptive field, it is helpful to select multiple 3x3 convolution kernel instead of the large convolution kernel. This can reduce the network parameters and increase the non-linearity, making the model more discriminative. Then the deeper layer of convolution is needed to continue to learn more abstract characteristics. In order to avoid further reducing the resolution and excessive loss of information in the original image, the pooling operation is not continued [22]. Finally, 4-time upsampling is done to obtain the predicted intensity value of each pixel. In particular, pictures collected on the production line are directly used as model input without any preprocessing.

### 2.3 Traditional Methods to Equivalent ConvNet

Traditional methods are used to discretize continuous pixel-wise prediction of the segmentation network and eliminate the false detection as far as possible. The entire procedure integrates density slicing, region segmentation and area filtration.

Here, we transform several traditional methods and construct a ConvNet as our feature refining network. The refining section and connection relationships can be seen in Figure 4. It should be noted that all the operators in this article are pixel-wise operators and differentiable.

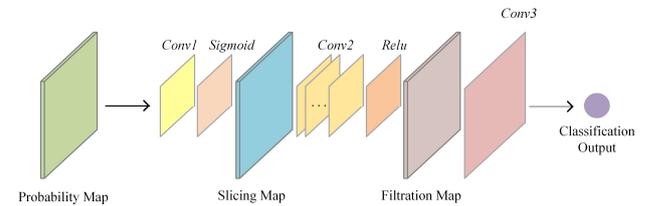

**Figure 4. Structure of the feature refining network.**

#### 2.3.1 Density Slicing

A pixel-wise method [23] discretizes intensity values of each pixel obtained by the segmentation network,

$$I^b(x,y) = \begin{cases} 255 & if \sum_{(i,j)\in R_s} W_{Gauss} * I^p(x_i, y_j) \geq T_1, \\ 0 & otherwise \end{cases} \quad (1)$$

Where $I(x,y)$ is the intensity value at pixel $(x,y)$ in segmentation output, $T_1$ is the intensity threshold for discretization, $*$ indicates a convolutional operator, $R_s$ is the eight close region of pixel $(x,y)$ used for calculating mean intensity values, $W_{Gauss}$ is Gaussian filter.

Because the strict binarization function is not differentiable, it cannot be implemented by a convolutional operation. Instead, among common activation functions, the sigmoid function can be approximately equivalent to a piecewise function of 0 and 1 with a threshold of 0. Before the sigmoid, we use a hand-marked convolution kernel to extract the intensity value of the corresponding pixel in input and set bias to divide these values into positive and negative categories. We adopt a gaussian filtering kernel with a size of 3x3 as the convolution kernel, and the bias is equivalent to the binarization threshold for training. It can be computed as,

$$I^b(x,y) = sigmoid(\sum w_i x_i - b_i), \quad (2)$$

This layer takes the results of the segmentation network as input and outputs the binary Map. This slicing operation can be regarded as a nonlinear pixel-wise accelerated discrete layer of the segmentation network.

### 2.3.2 Region Segmentation

Region segmentation, aimed to break weak connection between adjacent defect targets, can be implemented by the following operation as,

$$\begin{aligned} E(x,y) &= (I \ominus T)(x,y) = \underset{i,j=0}{\overset{m}{AND}}[I(x+i, y+i) \& T(i,j)], \\ D(x,y) &= (I \oplus T)(x,y) = \underset{i,j=0}{\overset{m}{OR}}[I(x+i, y+i) \& T(i,j)], \end{aligned} \quad (3)$$

Where $E(x,y)$ and $D(x,y)$ are the result of erosion and dilation respectively, $T(i,j)$ represents the structural element of the operation and its size is 3x3.

It is actually a shallow edge processing layer, which helps to separate the defects that are not tightly connected. These detached pixels can be eliminated in further steps. This operation can be equivalent to an area statistical layer and a segmentation layer and the detailed process can be seen in Section 2.3.3.

### 2.3.3 Area Filtration

The objective of Area filtration is to remove these isolated false detected points using two traditional steps. Firstly, a gradient-based direction detection method is used to detect the target edge and extract the contour area information can be transformed as,

$$\begin{aligned} P &= \nabla G_x * I(x,y) \\ Q &= \nabla G_y * I(x,y) \\ M[x,y] &= \sqrt{P[x,y]^2 + Q[x,y]^2}, \\ \theta[x,y] &= \arctan(Q[x,y]/P[x,y]), \end{aligned} \quad (4)$$

Where $P$ and $Q$ are the x− and y− gradients computed through Gaussian operators $\nabla G_x$ and $\nabla G_y$, $M$ is the gradient magnitude and $\theta$ is the output gradient orientation. It is actually a shallow ConvNet with 2 handcrafted kernels and a few merge layers.

Then, one learning-based segmentation method is used to train thresholds based on handcrafted features such as area of detected regions. This method can be computed as,

$$I^b(x,y) = \begin{cases} I^m(x,y) & if \sum_{(i,j)\in R_s} W_s * I^m(x_i, y_j) \geq T_2, \\ 0 & otherwise \end{cases} \quad (5)$$

Where $W_s$ is weights used for calculating the defect area near pixel $(x,y)$, $T_2$ indicates the intensity threshold for discretization, which can also be learned. This step, by setting the area threshold, can remove the small detection objects which have been disconnected in Section 2.3.2, and further reduce the false positive rate.

Then, we will discuss how to transform these operations into equivalent convolutional networks. As Region Segmentation and Area Filtration can both be equal to the combination of an area statistical layer and a segmentation layer, we integrate these two methods. To realize the area statistical layer, the defect area can be calculated by counting the number of pixels in the defect region. More specifically, we add up all defect pixels around $(x,y)$ using handcrafted convolution kernels which have a size of 9x9 and weights of all 1. Then in the segmentation layer, these relatively small intensity values can be divided to two parts by biases which is equal to the area-screening threshold. Finally, these low values will be eliminated through a ReLU function. It can be calculated as,

$$I^a(x,y) = relu(\sum ... \sum (w_j x_j - b_j)) \quad (6)$$

Where $I^a(x,y)$ represents the intensity value at pixel $(x,y)$ after area-screening operation, $b_j$ is the area-screening threshold, $\sum \cdots \sum$ means several consecutive convolutional operations.

A convolution is equivalent to a corrosion operation in morphology. Multiple convolutions will eventually filter out these isolated defects. This operation can be regarded as a nonlinear filtering layer of PartsNet, due to the fact that small-area false detection can often cause great interference to test results. Finally, the classification output is obtained through a full 1 convolution layer of image size.

### 2.4 Loss and Training

The loss cluster is shown in Figure 5. The total loss L is a weighted sum of 4 different losses and there are only 2 different types of losses.

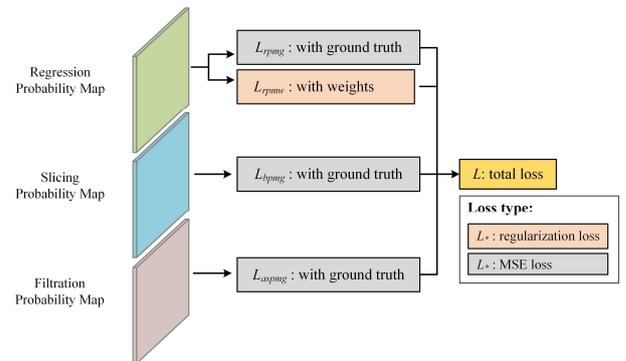

**Figure 5. The loss cluster illustrating our total loss. It is a weighted sum of 4 different losses coming from segmentation network, density slicing and area filtration.**

The loss function of the segmentation network is composed of MSE mean square error and the regularization term. MSE calculates the average loss of all pixel values. Weight regularization can minimize the weight and reduce the complexity of the network, which is necessary to prevent the model from fitting.

The loss function is defined as,

$$L_* = \frac{1}{W \cdot H} \sum_{i=1}^{W \cdot H} (p_i(x,y) - p_{i*}(x,y))^2 + \frac{\lambda}{n} \sum_w w^2 \quad (7)$$

where W and H are width and height of images, $p_i(x,y)$ and $p_{i*}(x,y)$ are the probability values at $(x,y)$ in the label map and predicted map respectively, n is the batch size, $\lambda$ is the regularization term coefficient used to weigh the proportion of regularization and MSE.

In order to train discrete thresholds of binarization and area-screening operations, we simply adopt MSE as the loss function. It can be calculated as,

$$L_{bpmg} = \frac{1}{W \cdot H} \sum_{i=1}^{W \cdot H} (p_i(x,y) - p_{i*}(x,y)*255)^2 \quad (8)$$

$$L_{aspmg} = \frac{1}{W \cdot H} \sum_{i=1}^{W \cdot H} (p_i(x,y) - p_{i*}(x,y))^2 \quad (9)$$

After model construction and data preparation, we conduct a three-step training procedure. Firstly, let Segmentation Detection Network learn defect features by training with MSE and regularization losses. After running a few epochs, we add slicing loss to train our threshold $T_1$. Finally, after another a few epochs, area-filtration loss is included to train threshold $T_2$. The idea enables PartsNet to learn step by step. Adam optimizer is adopted in all our training process.

## 3. EXPERIMENTS

We compare our model performance with other algorithms on different defect datasets to test PartsNet's generalization ability. As can be seen from the following experiments, our united PartsNet can implement density slicing, region segmentation, area filtration and classification without any fine tuning. And it performs better than classification networks and some common networks for objective detection.

### 3.1 Database

So far, there are few common databases available for automotive engine precision parts. Our training data was collected from the production line of several different domestic industries, including about 5 types of defective images. The first and second types are side and inside views of parts with metal chips attached, respectively. These metal residues, mainly composed of continuous chip and crack-chip, come from the machining processes of parts. The third type is the elevation view of parts with cracks in the bottom ring. The final two types are top and side surface scratch. Each image is 448 × 448 pixels in size and 500 pixels per inch (ppi) with manual marked pixel-wise label. The average number of images with each defect is about 500. PartsNet is respectively trained on the database and remains the same in the following experiments.

Our test data consists of 100 defective images and 100 non-defective images for each type, which has the same data distribution with the train dataset. Examples of typical parts defects are shown in Figure 6.

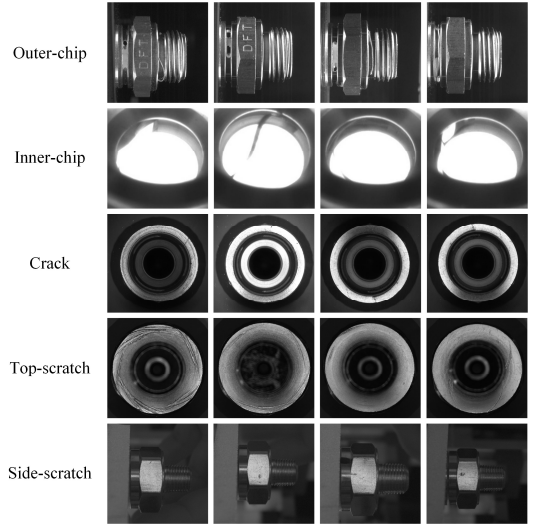

**Figure 6. Examples of database.**

### 3.2 Refining Network Performance

To evaluate the performance of our PartsNet, we test five different types of parts defects. Following [15], we compute pixel accuracy, mean accuracy and mean IU to measure the performance of defect segmentation. Results are shown in Table.1.

**Table.1. we compute pixel accuracy, mean accuracy and mean IU of five different parts defects, based on our proposed PartsNet with and without refining respectively**

| Types | Methods | Pixel acc. | Mean acc. | Mean IU |
|---|---|---|---|---|
| Outer-chip | Proposed | 0.9782 | 0.9474 | 0.9125 |
| | no refining | **0.9882** | **0.9631** | **0.9325** |
| Inner-chip | Proposed | 0.9584 | 0.8754 | 0.8476 |
| | no refining | **0.9607** | **0.9254** | **0.8804** |
| Crack | Proposed | 0.9825 | 0.8068 | 0.7808 |
| | no refining | **0.9925** | **0.8443** | **0.8344** |
| Top-scratch | Proposed | 0.9544 | 0.8733 | 0.8443 |
| | no refining | **0.9708** | **0.9122** | **0.8923** |
| Side-scratch | Proposed | 0.9624 | 0.8824 | 0.8812 |
| | no refining | **0.9751** | **0.9323** | **0.9267** |

For segmentation results, our integrated PartsNet gives a 0.64~1.27% accuracy improvement at pixel accuracy, 1.57~5% improvement at mean accuracy and 2~5.36% improvement at mean IU. The average accuracy improvement is 0.83%, 3.84% and 3.99% respectively. More test results are shown in Figure 9.

### 3.3 Classification Performance

The general objective way to confirm algorithm performance for classification is to calculate recall, precision, accuracy and F1-Score based on Confusion matrix.

Firstly, we draw Precision-Recall curve to evaluate the performance of defect extraction. Precision is defined as positive predictive value and recall is defined as the true positive rate.

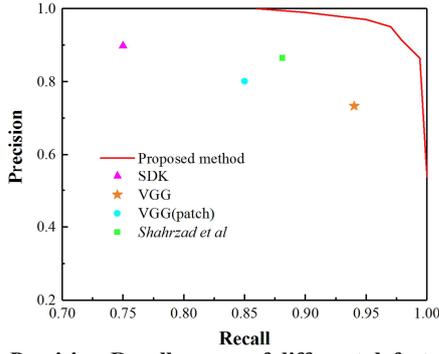

**Figure 7. Precision-Recall curves of different defect extraction algorithms on our test dataset.**

Figure 7 compares the defect extraction performance with other methods on our test dataset. The test dataset is composed of 200 images with crack. Traditional image processing algorithm is applied to extract defect feature, which includes a series of SDK functions. VGG is a typical classification Network and well used to recognize kinds of objects. VGG is re-trained and tested using the same dataset with us. To improve the performance of VGG on the defect database, images are then divided into subscale patches to boost significance of defect feature. *Shahrzad et.al.* constructed an effective network and it performed well on rail defects.

The best set of true positive rate and positive predictive rate are 98.2% and 91.1% respectively, because in industry the rate of missed inspection is required to be as low as possible. About 0.2 seconds is used on average to classify an image on our test dataset.

Then Table 2 computes F1-score and accuracy of different methods.

**Table 2. Algorithm performance of different methods based on the database of outer-chip defect**

| Methods | F1-score | accuracy |
| --- | --- | --- |
| Proposed | **47.2** | **97.5** |
| Proposed (no refining) | 45.6 | 90.5 |
| SDK | 39.8 | 76 |
| VGG | 41.1 | 74.3 |
| VGG (patch-wise) | 41.2 | 84.3 |
| *Shahrzad et.al.* | 42.8 | 86.5 |

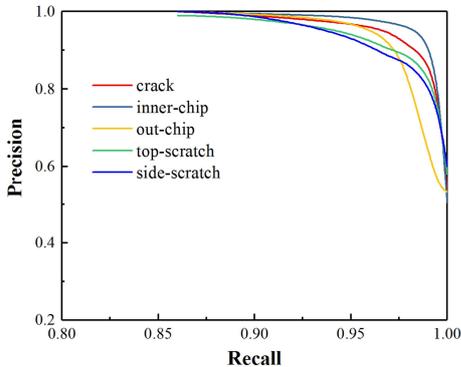

**Figure 8. Precision-Recall curves of different types of parts defects on our test dataset.**

Figure 8 compares extraction performance among different types of parts defects on our test dataset. The average precision and recall are 92.6% and 97.0% respectively.

Then Table 3 computes F1-score and accuracy of different types of parts defects based on our proposed PartsNet. The average mean F1-score and accuracy are 48.1% and 96.7% respectively.

**Table 3. Algorithm performance of different defects**

| Types of defect | F1-score | accuracy |
| --- | --- | --- |
| Outer-chip | 47.2 | 97.5 |
| Inner-chip | 49 | 98 |
| Crack | 48.3 | 96.6 |
| Top-scratch | 48.4 | 96 |
| Side-scratch | 47.3 | 95.5 |

## 4. CONCLUSION AND FUTURE WORK

In this paper, we propose a new method to construct a deep network called PartsNet for parts defect detection and network design. It combines the related knowledge of traditional feature processing and the representation ability of deep convolutional networks. Considering that the segmentation results are not accurate enough for classification prediction and may exist small-area false detection, we construct a refining network. The refining pipeline consisting of several typical traditional methods is equivalent to a simple network with empirical thresholds. The simple network is then transformed into a convolutional network and the thresholds are released to improve the adaptation ability and achieve end-to-end training. In this way, PartsNet outperforms other state-of-the-art open source detection algorithms and get reliable results on five different types of parts defects datasets.

Future work will include (1) exploring more high-efficient segmentation network structure, (2) optimization of networks and postprocess methods, and (3) extending PartsNet to more types of parts defects.

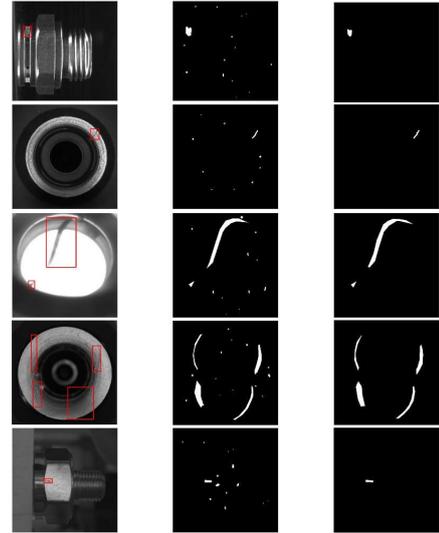

(a)defective images (b)segmentation map (c)refining results

**Figure 9. More results of our proposed PartsNet. Column (a)-(c) are defective images, segmentation map and refining results, respectively. From top to bottom, defects are sampled from our five different datasets acquired from industrial production lines. Red rectangles mark positions of defects.**